\documentclass[twocolumn,final,authoryear]{elsarticle}

\usepackage{ycviu}
\usepackage{framed,multirow}

\usepackage{amssymb}
\usepackage{latexsym}

\usepackage{times}
\usepackage{epsfig}
\usepackage{graphicx}
\usepackage{amsmath}
\usepackage{lineno,hyperref}

\newcommand{\comment}[1]{}
\usepackage{algorithm, algorithmic, ragged2e, booktabs,  bbm}

\usepackage{url}
\usepackage{xcolor}
\definecolor{newcolor}{rgb}{.8,.349,.1}

\usepackage{tikz}
\tikzstyle{state}=[shape=circle,draw=blue!50,fill=blue!20]
\tikzstyle{observation}=[shape=rectangle,draw=orange!50,fill=orange!20]
\tikzstyle{lightedge}=[<-,dotted]
\tikzstyle{mainstate}=[state,thick]
\tikzstyle{mainedge}=[<-,thick]

\usepackage{pgfplots}
\pgfplotsset{width=8.5cm,height=5cm}

\usepackage{subcaption, float, url}
\newcommand{\T}{{\scriptscriptstyle \top}}
\newcommand{\x}{\mathbf{x}}
\newcommand{\y}{\mathbf{y}}

\newcommand{\h}{\mathbf{h}}
\newcommand{\minisection}[1]{\vspace{1mm}\noindent{\textbf{\underline{#1}}:}}
\newcommand{\minisubsection}[1]{\vspace{1mm}{\textbf{#1}}}
\DeclareTextFontCommand{\empha}{\bfseries\em}

\journal{Computer Vision and Image Understanding}

\begin{document}
 \setcounter{page}{1}

\begin{frontmatter}

\title{CRF with Deep Class Embedding for Large Scale Classification}

\author[1,2]{Eran   \snm{Goldman}}
{\ead{eg4000@gmail.com}}
\author[1]{Jacob \snm{Goldberger}}
{\ead{jacob.goldberger@biu.ac.il}}

\address[1]{Faculty of Engineering, Bar Ilan University, Israel}
\address[2]{Trax Image Recognition}

\begin{abstract}
This paper presents a novel deep learning architecture for classifying structured objects in ultrafine-grained datasets, where classes may not be clearly distinguishable by their appearance but rather by their context. We model sequences of images as linear-chain CRFs, and jointly learn the parameters from both local-visual features and neighboring class information. The visual features are learned by convolutional layers, whereas class-structure information is reparametrized by factorizing the CRF pairwise potential matrix. This forms a context-based semantic similarity space, learned alongside the visual similarities, and dramatically increases the learning capacity of contextual information. This new parametrization, however, forms a highly nonlinear objective function which is challenging to optimize. To overcome this, we develop a novel surrogate likelihood which allows for a local likelihood approximation of the original CRF with integrated batch-normalization. This model overcomes the difficulties of existing CRF methods to learn the contextual relationships thoroughly when there is a large number of classes and the data is sparse. The performance of the proposed method is illustrated on a huge dataset that contains images of retail-store product displays, and shows significantly improved results compared to linear CRF parametrization, unnormalized likelihood optimization, and RNN modeling. We also show improved results on a standard OCR dataset.
\end{abstract}

\end{frontmatter}



\section{Introduction}

Object recognition is one of the fundamental problems in computer vision. It involves finding and identifying objects in images, and plays an important role in many real-world applications such as advanced driver assistance systems, military target detection, diagnosis with medical images, video surveillance, and identity recognition. Over the past few years deep convolutional neural networks (CNN) have led to remarkable progress in image classification
	\citep{ he2016deep,krizhevsky2012imagenet}, and resulted in reliable appearance-based detectors; e.g., \citep{lin2018focal,liu2016ssd,redmon2016yolo9000,ren2015faster,goldman2019dense}.

  Fine-grained object recognition aims to identify subcategory object classes, which includes finding subtle differences among visually similar subcategories such as dog breeds, product brands, car models, etc.
  The differences between classes are often small but always visually measurable, making visual recognition challenging but possible. Some of these datasets (e.g. UT-Zap50K \citep{finegrained}) provide each class with an \textit{in-vitro} image: a catalog or studio image isolated and captured under ideal imaging conditions; other datasets (e.g. Caltech-UCSD Birds \citep{WahCUB_200_2011}, Stanford Dogs \citep{KhoslaYaoJayadevaprakashFeiFei_FGVC2011}, FGVC-Aircraft
  \citep{maji2013fine}) provide each class with several \textit{in-situ} images, captured in natural real-world environments. Nonetheless, the image quality is mostly satisfactory for the task of visual classification. In fact, recent studies achieved good performance on fine-grained tasks
  \citep{Lin_2015_CVPR,peng2018object,Zhang2014}.

\begin{figure}[t]
\begin{center}
   \includegraphics[width=1.03\linewidth]{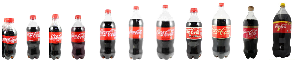}
\end{center}
   \caption{Spot the difference: Examples of classes with a similar appearance. Each product in this image belongs to a different category.}
\label{fig:classes}
\end{figure}

 However, the problem remains extremely difficult when the dataset categories are nearly identical in terms of their visual appearance. In this case, the object categories may be virtually indistinguishable, since the discriminant features are often masked by inadequate observation or visual artifacts. Here we present an \empha{ultrafine-grained structured classification dataset}; unlike other fine-grained classification datasets, our images are \textit{in-situ} low-resolution cropped patches whose classes are often virtually indistinguishable by visual inspection alone. Therefore, incorporating additional sources of information to the classifier is imperative. Since the object-patches originate from larger scenes, we can model contextual relations between the objects based on their geometric layout. This study tackles the challenge of fine-grained, large-scale structured classification, and describes a novel, state-of-the-art technique for this task.

\begin{figure}[h!!!]
\begin{center}
   \includegraphics[width=1\linewidth]{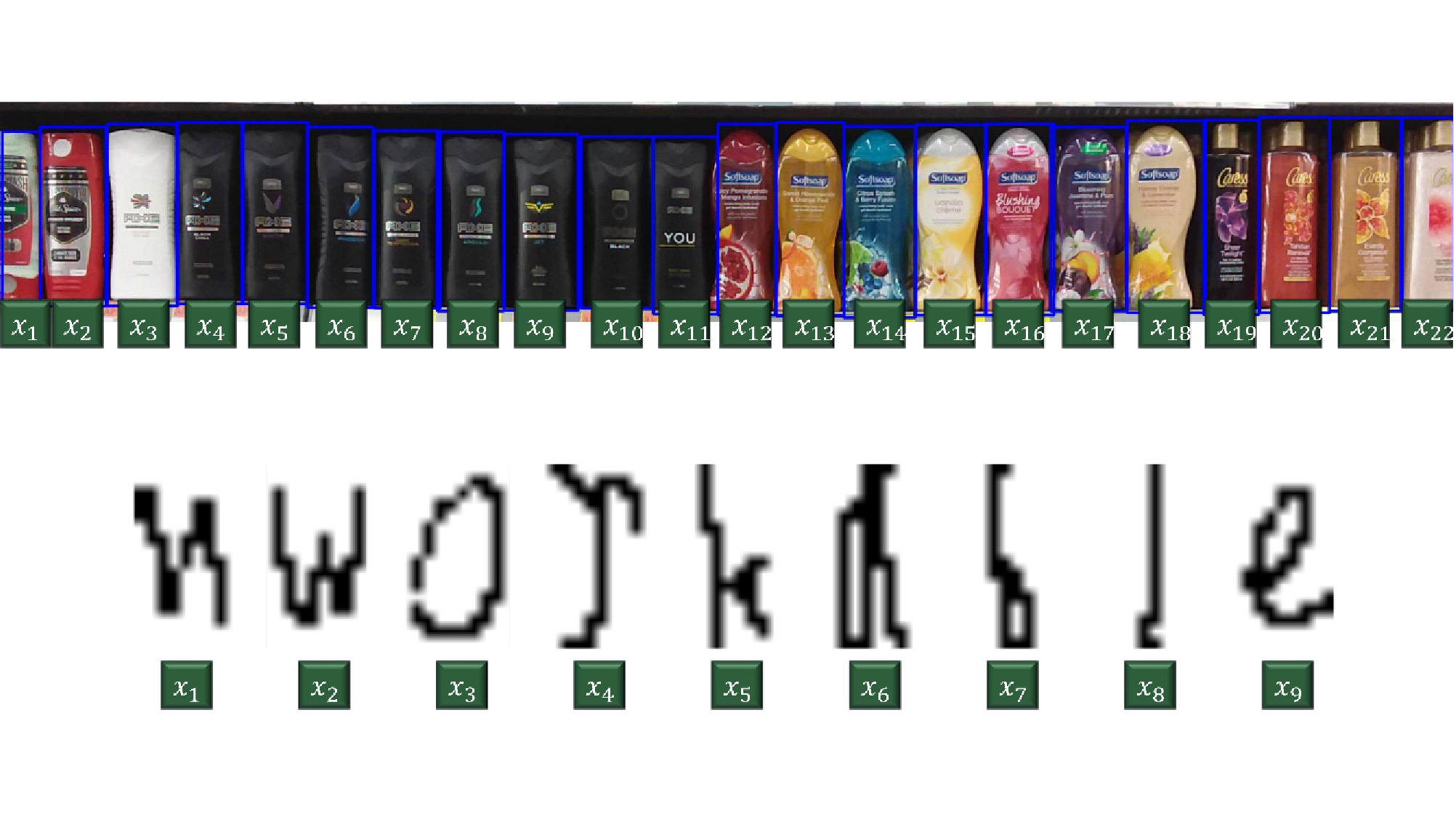}
\end{center}
\vspace{-8mm}
\caption{Examples of input sequences $\x$. The top sequence is from our store display dataset, and the bottom one is the word ``[U]nworkable" from the MIT OCR Handwritten Words Dataset which is  also used to validate our method.}
\label{fig:x}
\end{figure}

 We address the problem of classifying a sequence of objects based on their visual appearance and their relative locations.
Our dataset contains photos of retail store product displays, taken in varying settings and viewpoints. Our task is to identify the class of each product at the front of the shelves. The dataset is exclusively characterized by having a distinct geometric object structure made up of sequences of \textit{shelves}, a large number of classes, and very subtle visual differences between groups of classes in that some classes only differ in size or minor design details.
The unique challenges in this task involve (a) \textbf{large-scale classification}: handling the large number of possible classes, and (b) \textbf{ultrafine-grained structured classification}: the fact that the classes are not clearly distinguishable by their appearance but rather by their context. For example, products with an identical appearance but different container volumes are considered different classes (see examples in Fig. \ref{fig:classes}).

 Because an object's local appearance may not suffice for accurate categorization, additional information needs to be considered. In real world images, contextual data provides useful information about the spatial and semantic relationships between objects. Modeling a joint visual-contextual classifier is nontrivial in that some contextual cues are very informative, whereas others are irrelevant or even misleading
 \citep{barnea2019exploring,yu2016role}. Therefore, most deep learning detectors classify each detected object individually without taking the contextual information into account. Moreover, the handful of existing context-aware methods do not have the learning capacity for complex datasets such as ours, and cannot properly apply large-scale fine-grained structured classification.

 \minisection{Related Work} Context has been used to improve performance for image understanding tasks in various ways \citep{divvala2009empirical, felzenszwalb2010object,torralba2003contextual}. Graphical models have been widely applied to visual and auditory analysis tasks, by jointly modeling local features, and contextual relations. The tasks addressed by these models include image segmentation and object recognition \citep{Chandra, chen2016deeplab, gould2009decomposing,rabinovich2007objects,Peng,yao2012describing,zheng2015conditional}, as well as speech \citep{wang2012cocktail}, music \citep{korzeniowski2016fully}, text \citep{ chen2016word} and video analysis \citep{hu2014learning}.

 Few studies have applied deep learning features or detection results to context models:  \citet{chen2015learning} explored several techniques to learn structured models jointly with deep features that form MRF potentials.  \citet{chu2016deep} evaluated the performance of a joint CRF model on Faster R-CNN \citep{ren2015faster} detection results using an a-priori statistical summary for the pairwise potentials. \citet{korzeniowski2016fully} introduced a two-stage learning model for musical chord recognition: one network learns a single-frame representation, and the other learns the potentials of a linear-chain CRF model using the frame-representations as the CRF input.
  These models use the vanilla CRF parametrization, which includes pairwise potentials to represent object-pair interactions. They allocate a different parameter to each class pair. This approach, which ignores class similarities, is only sufficient for small sets of distinct classes. In effect, they have solely been tested on OCR datasets, which contain 26 classes \citep{chen2015learning}, a chord-recognition dataset with 25 classes \citep{korzeniowski2016fully} and PASCAL VOC 2007 with 20 classes \citep{chu2016deep}. However, this formulation is not sufficient for a large class-set that contains visually similar classes. Our dataset, which includes many visually similar categories, nearly a thousand classes and a million possible pairwise transitions overall, requires a more advanced learning mechanism. Furthermore, whereas in most previous object recognition studies the visual information was dominant, in our task, context information also makes a significant contribution.

In this study we present a Conditional Random Field (CRF) based method that explicitly learns the embedding of classes with respect to their neighbor's class and appearance. This is achieved by factorizing the CRF pairwise potential matrix to impose the structure of class embedding in a low-dimensional space. Our model learns the factorized parameters, and yields a joint contextual-visual embedding of the classes. The factorization drastically increases the learning capacity of contextual information, but also forms a multi-modal likelihood function which is more challenging to optimize. To overcome this, we develop a local surrogate likelihood and apply the proper regularization required for convergence. To train the network, we introduce a pairwise softmax architecture that optimizes a local approximation of the likelihood. Since the global factorized loss function is not convex, we favor optimizing the approximate surrogate likelihood, which allows us to include batch-norm related regularization for the object samples, and achieve dramatic improvement not only in training time and model simplicity but also in terms of the overall performance of the trained model. At test time, dynamic programming techniques are used for efficient exact inference of the classes.
The contribution of this work is twofold:
\begin{enumerate}
\item Combining deep class embedding into a CRF formulation that enables handling datasets with a huge number of classes.
\item  An approximated-likelihood training procedure that is both computational efficient and, unlike exact CRF likelihood, enables us to
incorporate batch-normalization into the training procedure.
\end{enumerate}

We validate our method on a large image  dataset and on an OCR dataset. Direct comparison of our method
to most relevant previous work show superior performance.
The rest of the paper is organized as follows.
In section \ref{sec:method} we  describe a CRF model with a class embedding formulation and present the learning and inference algorithms. Section \ref{sec:exp} contains a detailed data description and comparative experimental results. Object embedding analysis is described in Section \ref{sec:embedding}  and the conclusions are given in Section \ref{sec:conclusions}.

\section{CRF With Normalized Class Embedding}
\label{sec:method}
\subsection{Model Formulation}
\label{Model-Formulation-subsection}

 We are given a sequence of observations $\x = (x_1,...,x_n)$. The  data can be  a sequence of image patches which correspond to an horizontal layout of objects (see example in Fig. \ref{fig:x}).
 The goal is to classify each object in the sequence to one of a predefined $k$ categories where $k$ is a large number.  A standard CNN can classify the object in each image patch individually, implicitly assuming independence between elements in the sequence.  In order to include context in the classification process, we model the sequence as a CRF.

 We first use a local CNN to obtain a non-linear representation of the input image.
  Similar to the concept of transfer-learning, we can discard the CNN softmax layer, and use the convolutional layers to compute the feature-vectors of the input images. For each image-patch $x_t$ we define the feature vector $h_t = h(x_t)$ as the activations of the last hidden fully-connected or global-average-pooling (GAP) layer \citep{bengio2013representation, lin2013network}, and use it as the CRF input observation feature vector. 

 Linear-chain Conditional Random Field (LC-CRF) \citep{CRF2001} is a type of discriminative undirected probabilistic graphical model, whose conditional distribution $p(\y|\x)$  obeys a conditional Markov property. The joint probability distribution of a linear-chain CRF is:
\begin{equation}
p(\y|\x) = \frac{1}{Z}\prod_{t=1}^{n}\varphi(y_t,x_t,y_{t-1})\label{CRF}
\end{equation}
where $\x=(x_1,...,x_n)$ is the input sequence, $\y=(y_1,...,y_n)$ is the corresponding sequence of the target labels, $\varphi$ is the model's potential function and $Z$ is the partition function defined as the global probability normalization over all possible sequence label-assignments of length $n$.
We further assume that the potential function is defined as a simple log-linear function of the model parameters:
\begin{equation}
\varphi(y_t,x_t,y_{t-1}) = {\exp( y_{t-1}^{\T} P y_t  +  {h(x_t)^{\T}} U {y_t} + b^{\T}y_t)}.\label{likelihood}
\end{equation}
The CRF model parameters are $P$, $U$ and $b$ where $P$ is a  $k\times k$  pairwise potential matrix that models the relation between consecutive labels, $U$ is a unary potential matrix and the vector $b$ is the label bias. Note that we use a one-hot encoding for the labels.

The rationale for using a deep representation $h(x_t)$ for the input images is clear: as introduced by   \citet{krizhevsky2012imagenet}, the immense complexity of the visual object recognition task requires a model with a very large learning capacity. Convolutional layers provide the structure required for learning visual features of the \empha{unary} input. We aim to craft a suitable structure to learn the \empha{pairwise} contextual relations as well.

 CRF was originally applied to language processing tasks such as Part of Speech (POS) tagging and  Named Entity Recognition (NER) \citep{CRF2001}.
In most applications of CRF to either language or image understanding, there are no more than a few dozen different classes.
Our dataset contains nearly a thousand classes and the pairwise potential matrix $P$  has therefore nearly a million parameters. In order to properly learn and generalize the massive variety of possible neighboring patterns, we  enforce a structure on the pairwise potential matrix: the goal is to learn neighboring-class embedding in a feature vector space. For this purpose, we define a low-dimensional decomposition of the pairwise potential matrix $P$ as the product of the left-side neighbor embedding matrix $R$ and the class embedding matrix $Q$:
\begin{equation}
 P = R^{\T} Q.\label{embedding}
\end{equation}
The columns of $Q$ are low-dimensional embeddings of the target classes, and the columns of $R$ are embeddings of the classes of the left-side object.
Assigning the matrix factorization (\ref{embedding}) to the CRF potential function (\ref{likelihood}) we get:
\begin{equation}
\varphi(y_t,h_t,y_{t-1}) = \exp ((Ry_{t-1})^{\T}  Q y_t + {h_t^{\T}} U {y_t} + b^{\T}{y_t})
\label{potential}.
\end{equation}

Given the values of the model parameters, dynamic programming algorithms can be used for efficient and exact inference. The Viterbi algorithm finds the most probable sequence label assignment, and the Forward-Backward algorithm extracts the marginal probability of each item by summation over all possible assignments \citep{sutton2006introduction}.
The computational  complexity of the forward-backward and Viterbi algorithms is quadratic in the number of classes. In the next section we show that the matrix factorization improves classification performance. Note that the matrix factorization also improves the computation complexity of the dynamic programming algorithms used for the classification procedure.
The factorization brings the complexity  down  to the number of classes multiplied by the factorization dimensionality. 

\subsection{Learning the Model's Parameters}
\label{Training-subsection}
In the training phase we assume the availability of $s$ labeled sequences $(\x_1,\y_1),...,(\x_s,\y_s)$.  The
likelihood function of the factorized CRF model defined above is:
\begin{equation}\label{LossSeq}
\mathcal{L}(R,Q,U,b)  = \sum_{i=1}^s\log p(\y_i|\h(\x_i))
\end{equation}
where $\h(\x_i)$ is the feature vectors of the sequence $\x_i$ and 
$i$ goes over the sequences in the training data.  
 The likelihood function can be maximized  by applying standard Stochastic Gradient (SG) based methods.
 Since the CRF underlying graph is loop-free, it is tractable to compute the likelihood function and its gradient  using the forward-backward algorithm \citep{sutton2006introduction}. 
In case there is no low-dimensionality constraint on $P$, the likelihood is a concave function of the model parameters $P$,
$U$ and $b$  \citep{sutton2006introduction} and the optimal parameter can be easily found. The factorization of  the pairwise potential matrix $P=R^{\T}Q$ causes the likelihood (\ref{LossSeq}) to be a non-concave function of the model parameters and therefore there is no guarantee that gradient methods will converge to the global maximum likelihood. Hence, there no theoretical reason to favor optimizing the exact likelihood over approximate local variants that have better generalization capabilities. 

We next propose a novel learning approach that is based on optimizing an approximated CRF objective function, that can be used as a surrogate likelihood.
It also allows incorporating batch-normalization into the training procedure.
In the next section we show that this method, which learns to balance the CRF features, significantly improves the classification performance. 

Our approximated objective is inspired by the  MEMM formulation \citep{MEMM2000}. Linear-chain CRFs were originally introduced as an improvement on the Maximum Entropy Markov model (MEMM), which is essentially a Markov model in which
the transition distributions are given by a logistic regression model.
 CRF and MEMM can be written with the same set of parameters. The main difference between CRFs and MEMMs is that a MEMM uses per-state exponential models
for the conditional probabilities of next-states given the
current-state, whereas the CRF has a single exponential model
for the joint probability of the entire sequence of labels
given the observation-sequence. The MEMM directed graphical modeling in our case is:
\begin{equation}
p(\y|h(\x)) = \prod_{t=1}^n p(y_t|h_t,y_{t-1})
\label{memm_obj}
\end{equation}
where
\begin{equation}
p(y_t|h_t,y_{t-1}) = \frac{1}{Z(t)}{\exp( y_{t-1}^{\T}R^{\T} Q y_t  +  {h_t^{\T}} U {y_t} + b^{\T}{y_t})}\label{transition-memm}.
\end{equation}
When applying MEMM for inference it suffers from the label bias problem \citep{ Kakade2002,CRF2001} which may lead to a drop in performance in some applications. Here, however, we propose applying the MEMM objective only as a local approximation to learn the parameter set of the linear-chain CRF model whereas the test time inference still uses a global normalization of CRF modeling and thus avoids the label bias problem.
 In the appendix we review standard likelihood approximation strategies for efficient CRF training and show that the training method we use in this study can be viewed as a simplified version of the piecewise-pseudolikelihood approximation \citep{sutton2007piecewise}.

Our objective function is, therefore, defined as the conditional probability of the current-object class, given the class of the left-side neighbor object:
\begin{equation}\label{Loss}
\mathcal{L}_{ \tiny {\mbox{MEMM}}} = \sum_{i=1}^s \sum_{t=1}^n  \log p(y_{i,t}|h_{i,t}, y_{i,t-1})
\end{equation}
where $i$ goes over the sequences and $t$ goes over the objects in the sequence, $h_{i,t}$ is the object CNN-based representation, $y_{i,t}$ is the true class label and $p()$ is as defined at (\ref{transition-memm}).  Fig. \ref{fig:surrogate} illustrates the difference between the exact CRF objective and the MEMM objective we use as a CRF  approximation.
\begin{figure}[h!!!]
\center
\begin{subfigure}{0.9\linewidth}
\begin{tikzpicture}[]
\node[state] (y1) at (0,2) {$y_1$};
\node[state] (y2) at (2,2) {$y_2$}
    edge [blue, very thick] (y1);
\node[state] (y3) at (4,2) {$y_3$}
    edge [blue, very thick] (y2);
\node[state] (y4) at (6,2) {$y_4$}
    edge [blue, very thick] (y3);
\node[observation] (x1) at (0,0) {$x_1$}
    edge [blue, very thick] (y1);
\node[observation] (x2) at (2,0) {$x_2$}
    edge [blue, very thick] (y2);
\node[observation] (x3) at (4,0) {$x_3$}
    edge [blue, very thick] (y3);
\node[observation] (x4) at (6,0) {$x_4$}
    edge [blue, very thick] (y4);
\end{tikzpicture}
\end{subfigure} 

\vspace{0.5cm}

\begin{subfigure}{0.9\linewidth}
\begin{tikzpicture}[]
\node[state] (y1) at (0,2) {$y_1$};
\node[state] (y2) at (2,2) {$y_2$}
    edge [<-, blue, very thick] (y1);
\node[state] (y3) at (4,2) {$y_3$}
    edge [<-, blue, very thick] (y2);
\node[state] (y4) at (6,2) {$y_4$}
    edge [<-, blue, very thick] (y3);
\node[observation] (x1) at (0,0) {$x_1$}
    edge [->, blue, very thick] (y1);
\node[observation] (x2) at (2,0) {$x_2$}
    edge [->,blue, very thick] (y2);
\node[observation] (x3) at (4,0) {$x_3$}
    edge [->,blue, very thick] (y3);
\node[observation] (x4) at (6,0) {$x_4$}
    edge [->,blue, very thick] (y4);
\end{tikzpicture}
\end{subfigure} 
\caption{Graphical models of the CRF (top) and the MEMM (bottom)  objective functions.}
\label{fig:surrogate}
\end{figure}

The proposed objective  (\ref{Loss}) 
does not necessarily eliminate significant contextual information. Rather, when learning the  factorized CRF, it may enrich the training dataset, improve the stochastic nature of the SG optimization process and help to prevent overfitting since there are many more object samples than sequence samples, and the mini-batches are composed of adjacent pairs of objects taken from random training samples. In contrast, restricting the mini-batches to contain full sequences, would decrease the model's freedom to discover better solutions for the objective of pairwise transition parameters. In fact, as we empirically show in the next section, optimizing (\ref{Loss})   yields better results than optimizing the exact likelihood (\ref{LossSeq}). Note that the computational complexity of the approximated likelihood (\ref{Loss}) is linear in the number of classes and therefore the training process is much faster.

\subsection{Feature Scaling with Batch Normalization}
\label{BN}
In optimization, feature standardization or whitening is a common procedure that has been shown to reduce the convergence rates
\citep{orr2003neural}. In deep neural networks, whitening the inputs to each layer may also prevent converging into poor local optima. However, training a deep neural network is complicated by the fact that the inputs to each layer are affected by the parameters of all preceding layers, and need to continuously adapt to the new distribution. The batch-normalization (BN)
\citep{ioffe2015batch} method draws its strength from making normalization part of the model architecture and performing the normalization for each training mini-batch.
\begin{figure}
\begin{center}
   \includegraphics[width=1.05\linewidth]{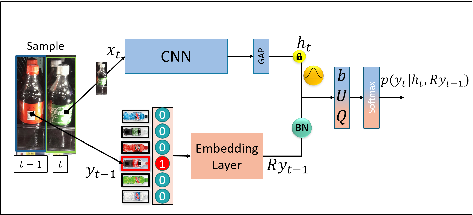}
\end{center}
  \caption{The \empha{approximate MEMM Likelihood} training architecture. An input sample is composed of an object and its left neighbor's label. The object image $x_t$ is converted to its deep visual representation $h_t$ by the CNN, whereas the label $y_{t-1}$ is converted to its deep second-order class representation by the embedding layer. The concatenated representation is the MEMM input.}
    \label{fig:pairwise_softmax}
\end{figure} 

 Our MEMM based training objective function 
 (\ref{Loss}) models the current state class  conditional distribution by a logistic regression (LR) model. Therefore, Eq. (\ref{transition-memm}) can be  written as:
\begin{equation}
\label{transition-memm-lr}
\log p(y_t|h_t,y_{t-1}) = \begin{pmatrix} 1  & h_t^{\T} &  (Ry_{t-1})^{\T}  \end{pmatrix}
\begin{pmatrix} b^{\T} \\ U \\ Q  \end{pmatrix} y_t - \log Z(t).
\end{equation}
  The first term is the input features to the LR and the second  term is the LR parameters  $Q$, $U$ and $b$. The input features  are composed of the  visual features of the CNN $h_t$ and the learned neighbor embeddings $Ry_{t-1}$. Since $y_{t-1}$  is a one-hot vector, the columns of $R$ are dense representations of the classes.  
 The standardization of the input feature vector $(Ry_{t-1},h_t)$ is important in order to avoid inherent bias between the local-visual and contextual class information. The goal is to encourage each input feature to have normalized mean and variance. Since the CNN is pre-trained we can compute the mean and variance vectors of the visual features as a pre-processing stage, and use standardized CNN feature vectors ${h}$ for CRF training and inference. In contrast, the context class embeddings $Ry_{t-1}$ are jointly learned with the LR  layer and thus are changed during the training process. Hence, we use the batch-normalization \citep{ioffe2015batch} method to learn their mini-batch normalization during the training process. 

 The input layer and the LR  layer of the MEMM training procedure are  illustrated in Fig. \ref{fig:pairwise_softmax}.

 Formally, by applying batch normalization to the context representation $Ry_{t-1}$, Eq. (\ref{transition-memm}) is replaced by:
\begin{equation}
p(y_t|h_t,y_{t-1}) = \frac{1}{Z(t)}{\exp( BN(Ry_{t-1})^{\T} Q y_t  +  {h_t^{\T}} U {y_t} + b^{\T}{y_t})}\label{transition-memm_bn}
\end{equation}
where
$\text{BN}{(x_i)} = \dfrac{x_i-\mu_\mathcal{B}}{\sqrt{\sigma_\mathcal{B}^2+\epsilon}}$
with mini-batch mean and variance:
$$\mu_\mathcal{B} = \dfrac{1}{m}\sum_{i=1}^m x_i \hspace{1cm} \sigma_\mathcal{B}^2 = \dfrac{1}{m}\sum_{i=1}^m (x_i-\mu_\mathcal{B})^2.$$
This way, the batch-normalization encourages the activations to have standard distributions during training, and tracks the moving averages of normalization parameters for the inference stage. The training method is summarized in Table \ref{alg_tbl}.

In the CRF exact likelihood, the weights in each sequence-level sample are shared across multiple locations in varying numbers of appearances, and the potential factors $R,Q$  are jointly used.  Hence, batch normalization cannot be applied directly to learn mini-batch statistics, and, as a matter of fact, previous CRF studies (e.g. \citep{Chandra,Peng, durrett2015neural}) did not use it.
A major advantage of the approximate MEMM likelihood we use is that the potential factors $ P = R^{\T} Q$ are used in a sequential order: We first apply the matrix $R$ on the context label and then the matrix $Q$ on the current label (see Eq. \ref{transition-memm-lr}). This is a natural setup for BN, which allows a very simple and effective way to apply batch-normalization to each neighboring-label sample.

\begin{table}[t!!!]
\normalsize
\caption{CRF with normalized deep class embedding algorithm}
\label{alg_tbl}
 \centerline{
 \fbox{\parbox{3.3in}{ 
\vspace{0.1cm}
{\bf Training Algorithm}:\\
Training data:  Image sequences $\x_1,...,\x_s$ with corresponding label sequences  $\y_1,...,\y_s$.\\
\begin{itemize}
\item Train a CNN to maximize the likelihood:
$$\mathcal{L}_{\mbox{cnn}} = \sum_{i=1}^s \sum_{t=1}^n  \log p(y_{i,t}|x_{i,t})$$
\item Apply SG to optimize the approximate likelihood:
$$\mathcal{L}_{ \tiny {\mbox{MEMM}}}(R,Q,U,b) = \sum_{i=1}^s \sum_{t=1}^n  \log p(y_{i,t}|h_{i,t}, y_{i,t\!-\!1})
$$
$$ \mbox{s.t.} \,\,\,  p(y_{i,t}|h_{i,t}, y_{i,t\!-\!1})  \propto \hspace{8cm} $$ $$
\hspace{0.2cm}  {\exp( BN(Ry_{i,t\!-\!1})^{\T} Q y_{i,t}  +  {h_{i,t}^{\T}} U {y_{i,t}} + b^{\T}{y_{i,t}})}
$$
and  $h_{i,t}$ is the CNN-based representation of $x_{i,t}$.
\end{itemize}
\vskip0.5cm 
{\bf Inference Algorithm}:\\
Input data: an image sequence $\x=(x_1,...,x_n)$.
\begin{itemize}
\item Apply CNN to obtain a representation $h_t=\h(x_t)$.
\item Apply the forward-backward (or Viterbi) algorithm on the following CRF:
$$
\hspace{-0.5cm} p(\y|\x) \propto
    \exp( \sum_{t=1}^n  BN_{test}(Ry_{t\!-\!1})^{\T} Q y_{t}  +  {h_{t}^{\T}} U {y_{t}} + b^{\T}{y_{t}})
$$
 to find the labels $\y=(y_1,...,y_n)$ of the object sequence.
\end{itemize}
}}}
\end{table}

After convergence of the training stage, we apply the batch-normalization inference procedure: Each column in matrix $R$ is standardized by the training population statistics $\mu_\mathcal{P}$ and $\sigma_\mathcal{P}$, estimated from the moving averages of mini-batch statistics tracked during training \citep{ioffe2015batch}:

$${BN_{test}}{(R{y_i})} = \dfrac{R{y_i}-\mu_\mathcal{P}}{\sqrt{\sigma_\mathcal{P}^2+\epsilon}}$$

At test time, we compute the standardized CNN representation vector ${h}$ for each object in the sequence, and classify the objects using the forward-backward algorithm as described above. The inference procedure is  summarized in Table \ref{alg_tbl}.

\section{Experiments}
\label{sec:exp}

\subsection{The Dataset}
Our dataset contains sequences of fixed-size image patches, originated from \textit{in-situ} photos of retail store displays taken in supermarkets and grocery stores. The objects are the inventory items positioned at the front of the displays, and the classes are their stock-keeping-unit (SKU) unique identifiers. Each object was originally annotated by its class label and bounding-box coordinates. The image patches were cropped and reshaped into single-object images of size $150 \times 450$ pixels, and grouped into shelves; i.e., sequences of horizontal layouts, arranged from left to right.
 The benchmark contains 76,081 sequences of 460,121 single-object images, originated from 24,024  photos of store displays. Each object is labeled as one of 972 different classes. Sequence lengths can vary from 2 to 32, and are typically between 4-12. The average sequence length is $6$ and the length standard deviation is $2.4$.
  To perform $k$-fold cross-validation, we split the dataset into 5 mixes of 80\% training and 20\% testing.

 Many groups of classes belong to the same archetype, and only differ in terms of minor details such as volume, flavor, nutrient content etc. They often share similar visual features, which makes appearance-based classification very difficult. On the other hand, the object layout behavior is very coherent: it is dictated by the supplier \textit{planograms} (specified product layouts) and extracted from the image \textit{realograms} (observed product layouts). Although realograms are non-deterministic by nature, consistent semantic patterns are frequently spotted. Class transition behavior may be discovered, revealing tendencies of pairs to appear as left-to-right neighbors, and individual classes to appear multiple times successively. The unique challenges we faced in our task are derived from the large number of visually similar classes, which co-occur in distinct structures in large-scale images.

\subsection{Implementation Details}
We first train a ResNet50 CNN \citep{he2016deep} from scratch to compute the hidden representation vector
  ${h}_{s \times 1}$ for each image-patch. In our implementation the hidden layer size (after global average pooling) was $s=2048$.
 Then, as a preprocessing step for the CRF model, we calculate the mean and standard deviation of each feature of the hidden representation vector from the training dataset: ${\mu}_{s \times 1}, {\sigma}_{s \times 1}$.

 The number of classes in our dataset is  $m=972$, and the class embedding dimensionality we use is $d=32$. We learn a class embedding matrix $Q_{d \times m}$, a neighbor embedding matrix  $R_{d \times m}$, a unary potential matrix $U_{s \times m}$ and a bias vector ${b}_{m \times 1}$. We initialize the bias parameter to $0$ and the weight parameters with random Gaussian samples $\mathcal{N}(0, 0.01)$ for symmetry breaking. We train the network as described in Section 2, using SG with mini-batches of size 128, and maximizing the log-likelihood function (\ref{Loss}) with  and $l_2$ regularization factor $\lambda= 5\cdot10^{-4}$ for all network parameters. The training samples in each mini-batch are object-pairs selected randomly from the benchmark.

Runtimes were measured on the same machine using an Intel(R) Core(TM) i7-5930K CPU @3.50GHz GeForce and a GTX Titan X GPU. A single batch epoch of the baseline unary system took 46 sec, a global optimization algorithm took 780 sec and our local optimization took  47 sec. The local training procedure is more efficient than computing the global maximum likelihood, because its time complexity is linear in the number of classes, whereas the global training procedure is quadratic in the number of classes. When using the factorized pairwise matrix, the global training time complexity can be reduced to the number of classes times the embedding dimensionality. The most important contribution of the approximate likelihood is hence in performance due to its  ability to add batch-normalization to the nonlinear objective.
 At test time it took less than 0.1 seconds to classify all the objects in a single image.

\subsection{Comparative experimental results}
In order to validate the performance of our method we implemented several alternatives.
They are all based on the same contextless CNN local information and only differ in the way they learn the object contextual information from the training dataset and integrate the context model with the local CNN. Below is a list of the baseline models we implemented.

\minisubsection{Unary} The baseline comparison model is the original CNN we trained without any contextual information.

\minisubsection{Pairwise Statistics}
Based on the work of \citet{chu2016deep}, we created a CRF model with unary potentials taken from the CNN classifier prediction results, and the pairwise potentials are  pairwise statistic  $P_{ij}=p(j|i) = p(y_t=j|y_{t-1} = i)$ that are estimated from the training dataset. In other words, the context information is modeled by a stationary
first-order Markov chain. No additional NN training is applied. The only single parameter we need to set is the relative weight of the unary and pairwise potentials.  This weight, which adjusts the tradeoff between the local appearance and the contextual information, was selected via cross-validation.

\minisubsection{Recurrent Neural Network}
Another modeling option for a sequence estimation is the Bi-Directional Recurrent Neural Network \citep{schuster1997bidirectional} with LSTM \citep{hochreiter1997long} as memory block (BiLSTM).
In this approach we compute the posterior distribution of the current object label based on all the visual information provided by the CNN: $p(y_t|\x) = p(y_t|x_1,...,x_n)$. The BiLSTM architecture learns a context vector $c_t$ for each object, which encapsulates the bidirectional information in the sequence input observations transferred from the CNN output $h_1,...,h_n$, and learns a softmax prediction $p(y_t|c_t)$ for each object label.
This approach, however, did not outperform the original contextless CNN.
The most important information, in addition to the object local appearance, is the label relations between neighboring objects which are not captured here; the BiLSTM network uses a softmax output layer that provides a separate prediction for each class and thus ignores class similarities. Although there is some structural similarity between RNN techniques and our local likelihood approach for CRF, the underlying probabilistic model is very different. Hence, CRFs are preferable for the element-wise classification of observed sequences because (a) they can explicitly learn second-order class similarity which is often the dominant source of contextual information, and (b) the Markovity assumption provides an optimal solver over the entire sequence.

\minisubsection{Log-linear CRF}
This method learns the log-linear parameters of the linear-chain CRF (\ref{likelihood}).
We implemented both the exact and  approximate likelihood training methods and tried both $l_1$ and $l_2$ regularizations for the pairwise potential matrix. We also tried whitening the one-hot input vectors. The results provided minor improvement over the baseline contextless classifier.

\minisubsection{Factorized CRF}
This method learns the factorized parameters of the pairwise weight matrix as defined in Eq. (\ref{potential}).

\minisubsection{Approximate Factorized CRF  + BN}
This is the model proposed in this study: The CRF pairwise weight matrix is factorized as defined in Eq. (\ref{potential}), the network is trained as described in subsection \ref{Training-subsection}, using the surrogate likelihood (\ref{Loss}), with $l_2$ weight regularization, and batch-normalization for the embedding features.

\begin{table}[th]
\normalsize
\caption{Comparison of the object-level error rate between the different methods and our full approach: \empha{Approximate Factorized CRF with BN}. The table shows the means and standard deviations of the error percentage over the 5 dataset splits. }
\begin{center}
\small
\begin{tabular}{|l|l|c|c|}
\hline
Method 	& Training & $\% \mu_{error} $ & $\% \sigma_{error} $ 	\\ \hline
Unary (no context)   &     CNN   	& 15.61 & 0.21          \\ \hline
BiLSTM               &      RNN     & 15.54 & 0.45          \\ \hline
Pairwise Statistics CRF &    CV    & 15.60 & 0.59   \\ \hline
Log-linear CRF & CRF 						 & 15.39 & 0.20 	\\ \hline
Log-linear CRF & MEMM 						 & 14.30 & 0.09 	\\ \hline
Factorized CRF	& CRF  				 & 14.62 & 0.19 	\\ \hline
Factorized CRF & MEMM 	 				 & 14.93 & 0.52 	\\ \hline
Factorized CRF  + BN & MEMM 			&  \bf{12.85} & 0.31  \\ \hline
\end{tabular}%
\end{center}
\label{table:results1}
\end{table}

\begin{figure}[]
\center
\begin{subfigure}{1\linewidth}
\begin{tikzpicture}[]
\begin{axis}[
    xlabel={Recall},
    ylabel={Precision},
    xmin=0.7, xmax=0.88,
    ymin=0.85, ymax=0.95,
    legend style={at={(0.26,0.03)},anchor=south, nodes={scale=0.5}},
    ymajorgrids=true,
    grid style=dashed,
    grid=major,
]
 
\addplot[
    color=blue,
    mark=square,
    ]
    coordinates {
    (0.8468,0.8468)(0.8467,0.8468)(0.8465,0.8487)(0.8455,0.8535)(0.8427,0.8595)(0.8382,0.8667)(0.8323,0.8741)(0.824,0.8826)(0.8139,0.8917)(0.7997,0.9014)(0.7818,0.9121)(0.7587,0.9249)(0.7336,0.935)(0.7058,0.945)(0.6737,0.9539)(0.6347,0.9616)(0.5877,0.9698)(0.5266,0.9763)(0.4418,0.9833)(0.3044,0.9902)
    };
    \addlegendentry{Unary}
    
\addplot[
    color=darkgray,
    mark=oplus,
    ]
    coordinates {
    (0.8478,0.8478)(0.8478,0.8478)(0.8476,0.8496)(0.8465,0.8538)(0.8444,0.8596)(0.8404,0.8665)(0.8349,0.8744)(0.8272,0.8827)(0.8175,0.8918)(0.805,0.9017)(0.7872,0.913)(0.7645,0.9253)(0.7411,0.9362)(0.7129,0.9455)(0.6813,0.954)(0.6443,0.9618)(0.5991,0.9693)(0.5397,0.9761)(0.4556,0.9825)(0.3197,0.988)
    };
    \addlegendentry{Global-Linear}

\addplot[
    color=violet,
    mark=triangle,
    ]
    coordinates {
    (0.8559,0.8559)(0.8559,0.856)(0.8558,0.857)(0.855,0.8601)(0.8533,0.8646)(0.8503,0.8703)(0.8458,0.8771)(0.8397,0.8848)(0.831,0.8928)(0.8199,0.9018)(0.8047,0.9122)(0.7848,0.9245)(0.7625,0.9354)(0.7366,0.9448)(0.7083,0.9531)(0.674,0.9607)(0.6326,0.9684)(0.5782,0.9753)(0.4989,0.9825)(0.3673,0.9885)
    };
    \addlegendentry{Global-Factorized}
 
\addplot[
    color=olive,
    mark=square,
    ]
    coordinates {
    (0.8585,0.8585)(0.8585,0.8585)(0.8584,0.8596)(0.8576,0.8626)(0.856,0.8673)(0.853,0.8733)(0.8486,0.8798)(0.8429,0.887)(0.8349,0.8944)(0.8246,0.9026)(0.8103,0.9128)(0.7922,0.9239)(0.7712,0.9336)(0.7486,0.9423)(0.7221,0.9513)(0.6898,0.959)(0.6503,0.9662)(0.5982,0.9735)(0.5246,0.9804)(0.4005,0.9876)
    };
    \addlegendentry{Approximate-Linear}
    
\addplot[
    color=cyan,
    mark=star,
    ]
    coordinates {
    (0.8435,0.8435)(0.8435,0.8435)(0.8433,0.8451)(0.8422,0.8492)(0.8398,0.855)(0.8362,0.862)(0.8307,0.8691)(0.8236,0.8779)(0.8138,0.8864)(0.8008,0.8959)(0.7839,0.9078)(0.7625,0.9202)(0.739,0.9316)(0.7122,0.9413)(0.6796,0.9502)(0.6445,0.9581)(0.6006,0.966)(0.5432,0.9732)(0.4614,0.9804)(0.3293,0.9873)
    };
    \addlegendentry{Approximate-Factorized}

\addplot[
    color=red,
    mark=+,
    ]
    coordinates {
    (0.8756,0.8756)(0.8756,0.8757)(0.8755,0.8762)(0.8751,0.8779)(0.8743,0.8806)(0.8729,0.8838)(0.8707,0.8873)(0.8677,0.8913)(0.8634,0.8961)(0.8575,0.9013)(0.8491,0.9074)(0.8376,0.9155)(0.8245,0.9231)(0.8103,0.9299)(0.7936,0.9366)(0.7748,0.9433)(0.7518,0.9499)(0.7209,0.9562)(0.675,0.963)(0.5898,0.9713)
    };
    \addlegendentry{Approximate-Factorized with BN}

\end{axis}
\end{tikzpicture}
\end{subfigure} 
\caption{Precision-Recall curves for the different training methods, obtained by sampling 20 different       confidence thresholds between 0 and 1.}
\label{fig:PR_curve}
\end{figure}

\minisubsection{Quantitative Results:}
Table \ref{table:results1} lists the results in terms of model error rate, indicates the incremental improvement in accuracy over model variations, and shows that the non-linear method based on batched-normalized class embedding yields significantly better results than the other alternatives.
Figure \ref{fig:PR_curve} depicts the Precision-Recall curve measured for the different methods by applying different confidence thresholds. It is particularly useful for our original objective, which aimed to maximize recall while preserving high (90\%+) precision. It shows that the MEMM based training method with batch normalization achieved significantly higher recall than the alternatives. Our method achieved a recall of $85.75\%$, whereas the unary baseline recall was $79.97\%$, and the second best alternative was $82.46\%$ - all preserving the same $90\%$ precision. It is worth pointing out that our test set is considerably large      which means that we correctly identified around 3,000 more objects than the unary model, and 2,000 more objects than the second best alternative.

\subsection{Ablation Study}

We performed an ablation analysis aimed at isolating the effect of the various innovations suggested. Each experiment uses the same configuration as in our method with only one alteration.

\minisubsection{Feature Scaling}
We tested the following variants:  removing the batch-normalization layer entirely,
removing the whitening of the CNN activations and Whitening the one-hot vectors at the input of the embedding layer instead of batch-normalizing its output.
In each case the results became much worse and were comparable to the contextless unary network.
On the other hand, when adding or removing the scale and shift from the BN parameters, the results remained comparable to our state-of-the-art results. This suggests that the BN layer has enormous impact since it whitens the embedding activations during training, similar to the whitening applied for the CNN activations.

\minisubsection{Regularization}
We tested $l_1$, $l_2$ or no regularization for the embedding weights. The results were significantly better with $l_2$ regularization, which encourages all the weights for each class in the embedding space to be used in training.

\minisubsection{Pairwise Matrix Factorization}
We considered other variants of the class embedding concept in which the embedding parameters of the target and neighboring labels are tied. For that purpose, we impose the structure of the embedding matrix $R$ on the current class as well as the neighboring class. The pairwise potential in this case is factorized as $P = R^{\T} D R$ to get the same embedding for the class and its neighbor. We may also apply the class embedding on the unary potentials matrix by factorizing $U = V^{\T} R$. In these parametrizations, applying the embedding-batch-norm requires parameter tying between the softmax inputs and the softmax weights, and thus compromises the effectiveness of the batch normalization process.

\minisubsection{Surrogate Likelihood Variances}
Global optimization of LC-CRF is not only much more time-consuming, but also lacks the ability to apply a straightforward batch normalization strategy, since the activations are shared in multiple locations in each sample in the mini-batch.
The same problem appears in other known methods of local likelihood approximation such as piecewise, pseudolikelihood and Piecewise-Pseudolikelihood (PWPL)  which are close variants of our local training model (See details in appendix \ref{local-likelihood}). Applying an embedding-batch-norm to the pseudolikelihood or PWPL methods would once again require parameter tying between the softmax inputs and the softmax weights. However, the PWPL in our case can be reduced to the form of a forward term which is equivalent to the MEMM-like objective (\ref{transition-memm}) and an additive backwards term which is independent of the CRF input. Hence, the MEMM-like objective function is theoretically highly related to PWPL. In the appendix we explain how our surrogate likelihood can be viewed as a simplified form of the PWPL objective.

\minisubsection{Different CNNs}
We tested our model with various CNN architectures: ResNet50 showed minor improvement in unary CNN performance over Alexnet
\citep{krizhevsky2012imagenet} and VGG
\citep{simonyan2014very}, and comparable improvement when incorporating our context-aware methods. Note that local visual information is often insufficient in our data, hence adding context is very helpful even with very strong CNNs.

\minisubsection{Different data}
We cross-validated our methods on 5 different train-test splits and obtained comparable results and small variances for the different mixes. We also verified our method on an OCR dataset of handwritten words as we describe in \ref{subsection:ocr}.

\minisubsection{Increased Learning Capacity}
We tried increasing the model's non-linearity by adding another fully connected layer and nonlinear ReLU between the one-hot vector input and the fully connected embedding layer.  We also tried learning the embedding in a higher dimensional space. These enhancements, however, did not improve performance, and turned out to be redundant.

\minisubsection{Similarity Networks} An additional noteworthy approach to identifying visually similar classes involves using an architecture which receives multiple samples as training input and compares pairs of samples in order to better discriminate between classes based on their visual features. These methods include Siamese Network
\citep{chopra2005learning, bromley1994signature} and other variants e.g.  \citep{kim2019edge, hoffer2015deep, zagoruyko2015learning}.

\begin{figure}[h!!!]
\begin{center}
   \includegraphics[width=1\linewidth]{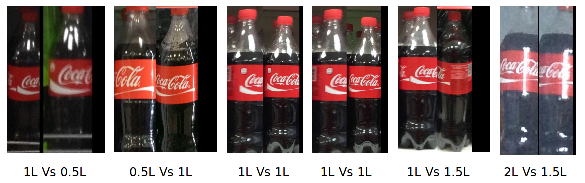}
\end{center}
   \caption{Examples of fixed-size input images which illustrate the futility of similarity based methods for our data.}
   \label{fig:perplexity}
\end{figure}

In our case, however, such approaches were unhelpful due to the limited priors and the ultrafine-grained nature of our dataset. For example, object volume or flavor are often visually unmeasurable. Furthermore, these approaches ignore class neighbor information which is usually the dominant source of contextual information in observed sequences.  Fig. \ref{fig:perplexity} exemplifies the type of difficulties we faced when trying to learn pairwise visual similarities in our data.

\begin{figure}[h!!!]
\begin{center}
   \includegraphics[width=1\linewidth]{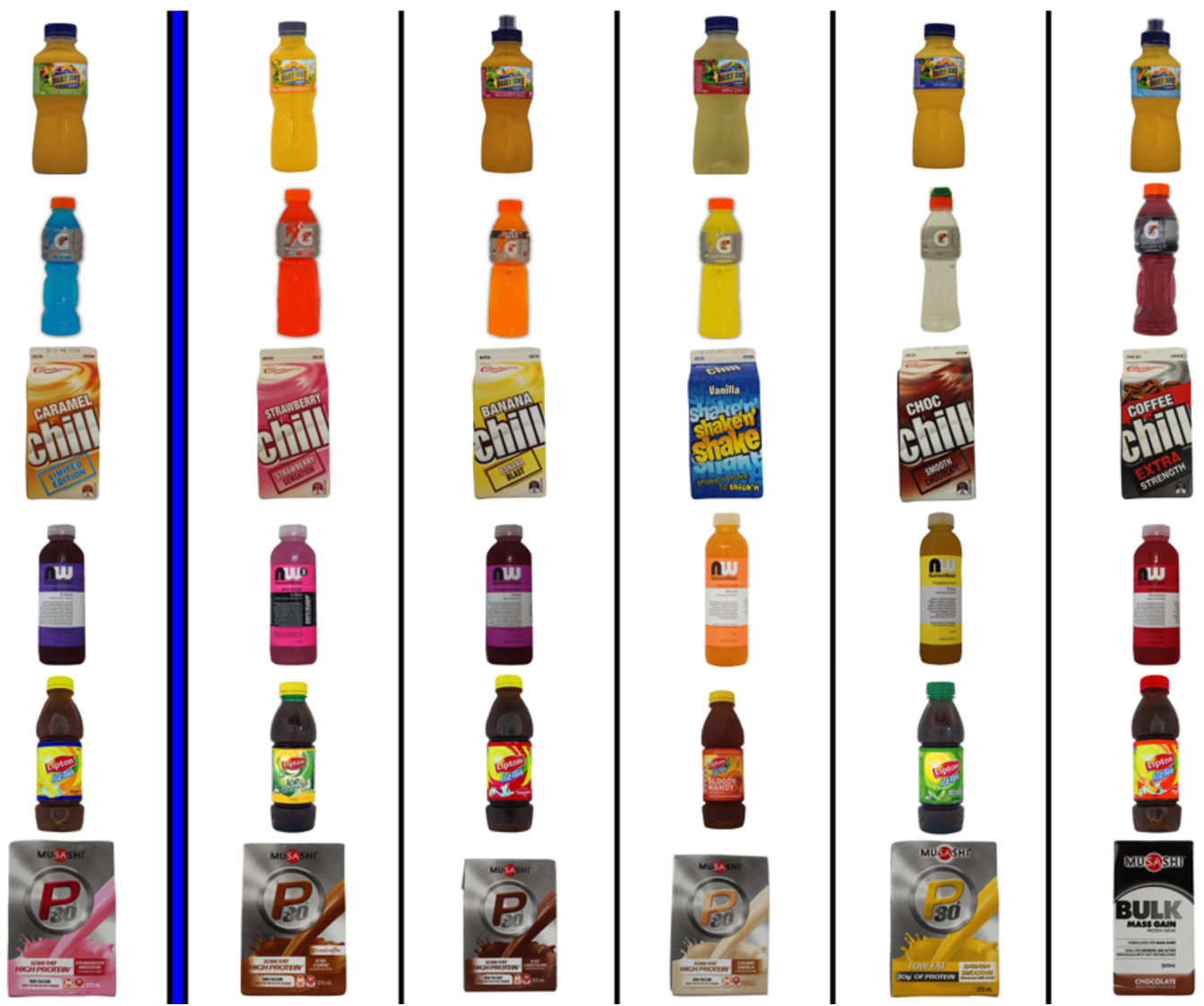}
\end{center}
   \caption{Class similarity examples. For each class we show five nearest neighbors based on the cosine distance computed on the class embeddings.}
   \label{fig:embedding_neighbors}
\end{figure}

\subsection{MIT OCR Dataset of Handwritten Words}  \label{subsection:ocr}
We tested our method on data from another benchmark, to validate that our approach generalizes well to other domains beyond store shelves and retail objects. Despite the absence of other fine-grained structured classification datasets of a similar scale, the \empha{MIT OCR Dataset of Handwritten Words} \citep{kasselocr} can provide a close approximation.
The dataset contains 6877 words, composed of 52152 samples of lowercase letters collected from 150 human subjects. Each sample is a $16 \times 8$ binary pixel image. We split the dataset into 5512 training words and 1365 testing words. The input features of the CRF are the original 128 pixels. There are only 26 classes in the dataset - the English alphabet letters. Therefore, we used a smaller embedding dimension of 16. We performed the following experiments: \empha{Unary}- prediction of the samples directly from the input without context, \empha{Log-linear CRF}- learning the log-linear parameters of the linear-chain CRF, \empha{Factorized CRF}- learning the factorized parameters of the linear-chain CRF,  and \empha{Factorized CRF + BN}- learning the factorized parameters while whitening the high level representation of the input image and neighbor embedding with BN. The CRF models were trained with the approximate likelihood architecture presented in this work. For each experiment, we report the accuracy (Acc), average precision (AP),  and the recall values attained for  precision of $0.7$ (R$^{P=0.7}$) and $0.9$ ( R$^{P=0.9}$). The results are shown in Table \ref{table:ocrresults}. Despite the small scale of the data and the small number of classes, our method yielded more accurate results than the alternatives. Especially, the batch normalization procedure applied to the context letter representation, was found to be beneficial.

\begin{table}[h!!]
\centering
\normalsize
\caption{Results on MIT OCR dataset of handwritten words.}
\begin{center}
\begin{tabular}{|l|c|c|c|c|}
\hline
Architecture 	    & Acc	& AP	& R$^{P=0.7}$	& R$^{P=0.9}$	 	\\ \hline
Unary         		& 0.74 	& 0.81 & 0.78 & 0.59          \\ \hline
Log-linear CRF 		& 0.81	& 0.88 & 0.87 & 0.73	\\ \hline
Factorized CRF	 	& 0.82	& 0.90 & 0.89 & 0.76 	\\ \hline
Factorized CRF + BN		& \bf{0.84} & \bf{0.92} & \bf{0.91} & \bf{0.79} \\ \hline
\end{tabular}%
\end{center}
\label{table:ocrresults}
\end{table}

\section{Class Embedding Analysis}
\label{sec:embedding}
As a byproduct of the classification model we also obtain a low-dimensional embedding of the different classes.
Each column of the neighbor embedding matrix $R$ is a vector representation of the corresponding class.
  A common similarity metric is the cosine of the angle between the vectors. We can measure the distance between classes by the cosine of their
  vector representation. Fig. \ref{fig:embedding_neighbors} shows several examples of an object class and its most similar classes.
 We can see that this similarity does not reflect visual appearance similarity, e.g. in the second example the similar classes have very different colors. This situation has been studied extensively for the linguistic problem of word embedding.
  The goal of word embedding algorithms is to represent similar words by similar vectors.
It is often useful to distinguish between two kinds of similarity or association between words \citep{Schutze}.
Two words are said to have first-order co-occurrence if they are typically nearby each other (e.g. wrote is a first-order associate
of book or poem). Two words have second-order co-occurrence if they have similar neighbors (e.g. wrote is a second-order
associate of words like said or remarked). Second-order word similarity is thus expected to capture a semantic meaning and
measure the extent to which the two classes are replaceable
based on their tendencies to appear in
similar contexts.
In  Fig. \ref{fig:embedding_neighbors}  we show that object class embedding captures second-order information. Proximity here corresponds to the mutual tendency to have similar neighbors. We can see in the figure that similar classes, although looking visually different, represent products of similar container-types, volumes and brands.

\begin{figure}[h!!]
\begin{center}
   \includegraphics[width=0.7\linewidth]{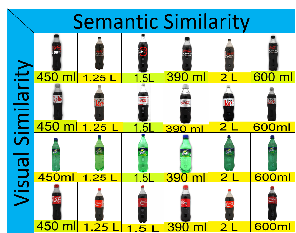}
\end{center}
   \caption{Examples of the two different types of similarities.}
   \label{fig:similarities}
\end{figure}

Visual similarity and second-order semantic similarity are based on two profoundly different criteria, and may be uncorrelated or even have a negative correlation in some cases as we demonstrate in Fig. \ref{fig:similarities}: classes are \textbf{visually close} when it is easy to confuse them based on their visual appearance, and are \textbf{semantically close} when it is statistically reasonable to switch one for the other on a shelf (i.e. ``synonymous" classes). The rows in Fig. \ref{fig:similarities} contain classes that are visually close but semantically far; i.e., they look alike but tend to appear in different contexts, whereas the columns contain classes which are semantically close but visually far; i.e., they look different, but tend to appear in similar contexts. The examples from the retail world refer to classes of similar brands but with different form-factors or volumes, which tend to appear in different displays in stores. A speech analogy would be comparing homophones (e.g. meet vs. meat, sale vs. sail) with synonyms (e.g big vs. large, fast vs. quick).

It is hence clear why these two types of similarity contribute two different types of information, and need to be used jointly for the task of object classification. The visual similarity is relevant for the visual image information whereas the class similarity in the embedded space is relevant for the contextual information.

\section{Conclusions}
\label{sec:conclusions}
We introduced a novel technique to learn deep contextual and visual features for fine-grained structured prediction of object categories, and tested it on a dataset that contains spatial sequences of objects, and a large number of visually similar classes.

 Our model clearly outperforms all the other tested models. This architecture appears to be the most straightforward generalization of a context-less classifier to become context-dependent when both the input and the context data require a large learning capacity: the network learns deep feature vectors for neighboring classes, analogously to the learned deep input representations. The  Markovity and stationarity assumptions make it sufficient to train with individual objects as samples to enrich the training data diversity, allow for simple embedding batch normalization, and boost the non-convex optimization process both in terms of time and performance.

\section{Appendices} \label{appendix}
\subsection{Local Likelihood Approximation}
\label{local-likelihood}
In this appendix we show how the objective function that we used for optimization is related to previously suggested approaches.
Pseudolikelihood \citep{besag1975statistical} is a classical approximation of the CRF likelihood function that simultaneously classifies each node given its neighbors in the graph. The pseudolikelihood objective function only depends on the object and its Markov blanket. The pseudolikelihood of our model (\ref{CRF}) is:
\begin{equation}
\label{pseudolikelihood}
\log p(\y|\x) = \sum_t \log p(y_t|y_{t-1}, y_{t+1},\x)
\end{equation}
where $p(y_t|y_{t-1}, y_{t+1},\x)$ is
\begin{small}
\begin{equation}
\frac{\exp({ y_{t-1}^{\T} P y_t  + y_{t}^{\T} P y_{t+1}+
 {x_t^{\T}} U {y_t} + {y_t^{\T}}b})}{\sum_a
 \exp({ y_{t-1}^{\T} P a  + a^{\T} P y_{t+1}+
 {x_t^{\T}} U {a} + {a^{\T}}b})}.
\end{equation}
\end{small}
Piecewise training \citep{suttonpiecewise} is a heuristic method to predict the graph factors from separate ``pieces" of the graph. The piecewise objective function is equivalent to the likelihood function of a node-split graph \citep{sutton2007piecewise}, which contains all the single-factor components split from the original graph. Using the CRF notation in Eq.  (\ref{CRF}), the piecewise likelihood approximation in our case is:
\begin{equation}
\label{piecewise}
\log p(\y|\x) = \sum_t \log \frac{\varphi(y_t,x_t,y_{t-1})}{\sum_{a,b}\varphi(a,x_t,b)}.
\end{equation}
Note that due to the term in the denominator, computing the piecewise likelihood is quadratic in the number of classes. Piecewise Pseudolikelihood (PWPL) is the standard pseudolikelihood applied to the node-split graph. Its computation is  efficient because the objective function is simply the sum of local conditional probabilities.  In our case, applying the pseudolikelihood approach on the piecewise objective (\ref{piecewise}) would give us the following PWPL form:
\begin{equation}
\log p(\y|\x) = \sum_t  \log ( \frac{\varphi(y_t,x_t,y_{t-1})}{\sum_{a}\varphi(a,x_t,y_{t-1})} \cdot \frac{\varphi(y_t,x_t,y_{t-1})}{\sum_{a}\varphi(y_t,x_t,a)} ).
\label{pwpl}
\end{equation}
   \citet{sutton2007piecewise} showed that in many cases the PWPL has better accuracy than standard pseudolikelihood, and in some scenarios has nearly equivalent performance to piecewise approximation and even to global maximum likelihood.
The first term  inside the $\log$ function is equivalent to the forward MEMM objective function (\ref{transition-memm}) that we used.
The second term can be written in the  form:
\begin{equation}
 p(y_{t-1}|y_t) = \frac{\exp( y_{t-1}^{\T} P y_t) }{\sum_a \exp( a^{\T} P y_t) }.\label{PWPL-backwards}
\end{equation}
This term  is independent of the CRF visual input.
The PWPL approximation can be thus expressed as:
\begin{equation}
\log p(\y|\x) = \sum_t ( \log p(y_t|y_{t-1}, x_t) + \log p(y_{t-1}| y_t) ).
\end{equation}

Hence the MEMM-like objective function we used (\ref{memm_obj})  can be viewed as a simplified version of the  piecewise-pseudolikelihood objective (\ref{pwpl}) that was found to be the preferred likelihood approximation  for language processing tasks \citep{sutton2007piecewise}.

\bibliographystyle{model2-names}
\bibliography{main}
\end{document}